\documentclass[letterpaper, 10 pt, conference]{ieeeconf}  

\IEEEoverridecommandlockouts                              
\usepackage[linesnumbered,ruled]{algorithm2e}
\usepackage{float}
\usepackage{subfigure}
\usepackage{graphics} 
\usepackage{epsfig} 
\usepackage{times} 
\usepackage{amsmath} 
\usepackage{amssymb}  
\usepackage{color}
\usepackage[table]{xcolor}
\usepackage[T1]{fontenc}
\usepackage{bm}
\usepackage{cite}
\usepackage{hyperref}
\usepackage{booktabs}
\hypersetup{
	colorlinks=true,
	linkcolor=blue,
	filecolor=blue,      
	urlcolor=blue,
	citecolor=cyan,
}
\usepackage{cite}
\title{\LARGE \bf BAANet: Learning Bi-directional Adaptive Attention Gates \\for Multispectral Pedestrian Detection
}

\author{Xiaoxiao Yang$^{1,3}$, Yeqiang Qian$^{2}$, Huijie Zhu$^{3}$, Chunxiang Wang$^{1}$, Ming Yang$^{1*}$
	\thanks{$^*$Corresponding author}
	\thanks{$^{1}$Department of Automation, Shanghai Jiao Tong University, Shanghai 200240, China, and also with the Key Laboratory of System Control and Information Processing, Ministry of Education of China, Shanghai, 200240, China. \tt\small xiaoxiao.yang@sjtu.edu.cn; \tt\small mingyang@sjtu.edu.cn }
	\thanks{$^{2}$University of Michigan-Shanghai Jiao Tong University Joint Institute, Shanghai Jiao Tong University, Shanghai, 200240, China. \tt\small qianyeqiang@sjtu.edu.cn}
	\thanks{$^{3}$  Science and Technology on Near-Surface Detection Laboratory, Wuxi, 214035, China. }
}

\begin{document}
	
	\maketitle
	\thispagestyle{empty}
	\pagestyle{empty}
	
	\begin{abstract}
 	Thermal infrared (TIR) image has proven effectiveness in providing temperature cues to the RGB features for multispectral pedestrian detection.
 	Most existing methods directly inject the TIR modality into the RGB-based framework or simply ensemble the results of two modalities.
 	This, however, could lead to inferior detection performance, as the RGB and TIR features generally have modality-specific noise, which might worsen the features along with the propagation of the network.
 	Therefore, this work proposes an effective and efficient cross-modality fusion module called Bi-directional Adaptive Attention Gate (BAA-Gate).
 	Based on the attention mechanism, the BAA-Gate is devised to distill the informative features and recalibrate the representations asymptotically.
 	Concretely, a bi-direction multi-stage fusion strategy is adopted to progressively optimize features of two modalities and retain their specificity during the propagation.
 	Moreover, an adaptive interaction of BAA-Gate is introduced by the illumination-based weighting strategy to adaptively adjust the recalibrating and aggregating strength in the BAA-Gate and enhance the robustness towards illumination changes.
 	Considerable experiments on the challenging KAIST dataset demonstrate the superior performance of our method with satisfactory speed. 
	\end{abstract}

	\section{INTRODUCTION}
	Nowadays, visible pedestrian detection has played a crucial and fundamental role in various computer vision applications, such as autonomous driving~\cite{Yang2018realtime, qian2018pedestrian}, video surveillance~\cite{bilal2016low}, robotics~\cite{aguilar2017pedestrian, 2019DiasICRA}, \textit{etc}.
	Except for the visible modality, the thermal infrared (TIR) modality is generally adopted as the complementary information for RGB-T detection, due to its capability of capturing thermal features in poor illumination conditions.
	The combination of two spectral can further enhance the detecting performance during the complex illumination environments, meaningful for the around-the-clock missions.
	Therefore, multispectral pedestrian detection has been extensively studied, particularly since the proposal of the KAIST multispectral pedestrian dataset~\cite{liu2016multispectral, choi2016multi, konig2017fully, guan2019fusion, zhang2020multispectral, li2019illumination, zhang2019weakly, kieu2020task}.
	
	\begin{figure}[]
		\includegraphics[width=0.5\textwidth]{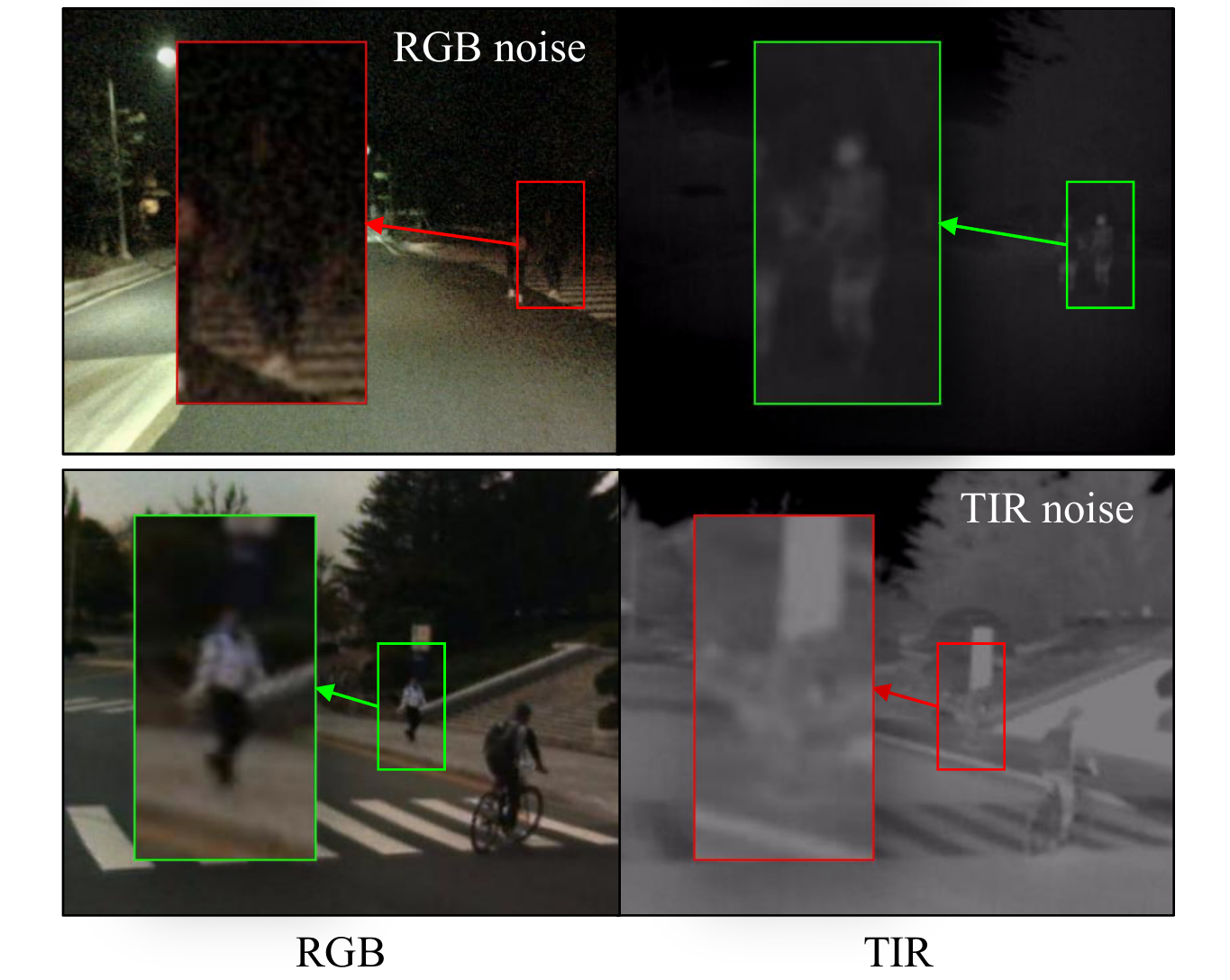}
		\caption{The illustration of modality-specific noise in two modalities.
		The first row indicates the RGB noise, which is presented as a similar object appearance in the poor illumination.
		The second row demonstrates the TIR noise manifests as objects with a similar temperature are prone to be mixed together.
		}
		\label{fig1}
	\end{figure}
	
	Although great progress has been made~\cite{cao2021handcrafted}, most existing methods directly inject the TIR information into the RGB-based framework or simply ensemble the results of two modalities.
	This, however, exists two significant challenges:
	(1): \textit{The essential difference between the RGB and TIR modalities.}
	The RGB image reflects color information, while the TIR image reflects temperature information. They show the different inherent properties of objects. 
	Therefore, how to identify the difference between modalities and unify them into an effective representation is still an open problem.
	(2): \textit{Modality-specific noise in the RGB and TIR modalities.}
	The RGB images are generally noisy in the poor illumination environment due to the limited lighting sensibility of the sensor, which is presented as the similar appearance of objects.
	Besides, the TIR-specific noise manifests as the objects with similar temperatures are prone to blend with surroundings.
	Meanwhile, TIR modality lacks texture details and precise edges due to the limited performance of infrared cameras.
	Fig.~\ref{fig1} illustrates the noise of two modalities.
	These noises in two modalities might be superimposed and worsen the robustness and accuracy, leading to detection failure.

	For the aforementioned problems, most existing approaches focus on tackling the first problem.	
	The standard fusion framework is to extract features of two modalities via two CNN branches, and then directly concatenate them or simply ensemble the results of two modalities.
	To identify the best stage to fuse features, \cite{wagner2016multispectral, liu2016multispectral} design different network architectures which concatenate features on different stages.
	\cite{zhang2019weakly} proposes to align the region features of two modalities by the Region Feature Alignment module to enhance the fusion quality.
	\cite{MBNetECCV2020} proposes to tackle the modality imbalance problem based on the differential modality aware fusion module.
	Despite the plausible solutions provided by these approaches, we argue that these methods exist intrinsic incapability due to the implicitly modeling of correlation between modalities, limiting the representation potential of fused features. 
	Meanwhile, the modality-specific noise in the second challenge is not considered, which might deteriorate features fusion as the network propagates forward.

	Based on the previous analysis, we propose an efficient and effective framework embedded with a novel cross-modality fusion gate called Bi-directional Adaptive Attention Gate (BAA-Gate). 
	Concretely, inspired by the attention mechanism, the proposed framework utilize the correlation of two modalities to first suppress the exceptional features in the TIR modality around similar temperature regions and focus on the informative object features.
	Then, the distilled TIR features are used to recalibrate the RGB modality further.
	Meanwhile, considering the RGB features are generally noisy, especially in poor illumination environments, the bi-directional multi-stage fusion strategy is adopted to suppress the undesired RGB features. The refined RGB features can also be utilized to complete the TIR modality. 
	In this way, the features of the two modalities are progressively distilled and recalibrated to enhance the quality of representations for the final detection.
	Besides, due to the confidence level of modalities is different in various illumination environments,  adaptive interaction of BAA-Gate is implemented by introducing the illumination-based weighting strategy to adaptively adjust the recalibration and aggregation strength between modalities and the contributions of two modalities for the detection result.
	The main contributions of this work are three-fold:
	\begin{itemize}
		\item 
		This work proposes a novel multispectral pedestrian detector called BAANet.
		With the proposed BAA-Gate, the network could distill and recalibrate features of two modalities by stages and enhance the discriminability of fused representations. 
		\item An adaptive interaction strategy based on illumination-based weighting is introduced to adaptively control the recalibration and aggregation strength between modalities and adjust their contributions to the final detection results.
		
		\item The presented BAANet is evaluated on the challenging KAIST multispectral pedestrian dataset.
		Compared with the other ten state-of-the-art approaches, the experiments show the competitive performance of the BAANet in terms of robustness and accuracy with satisfying speed.
		
	\end{itemize}

	\begin{figure*}[!h]
	\centering
	\includegraphics[width=1.04\textwidth]{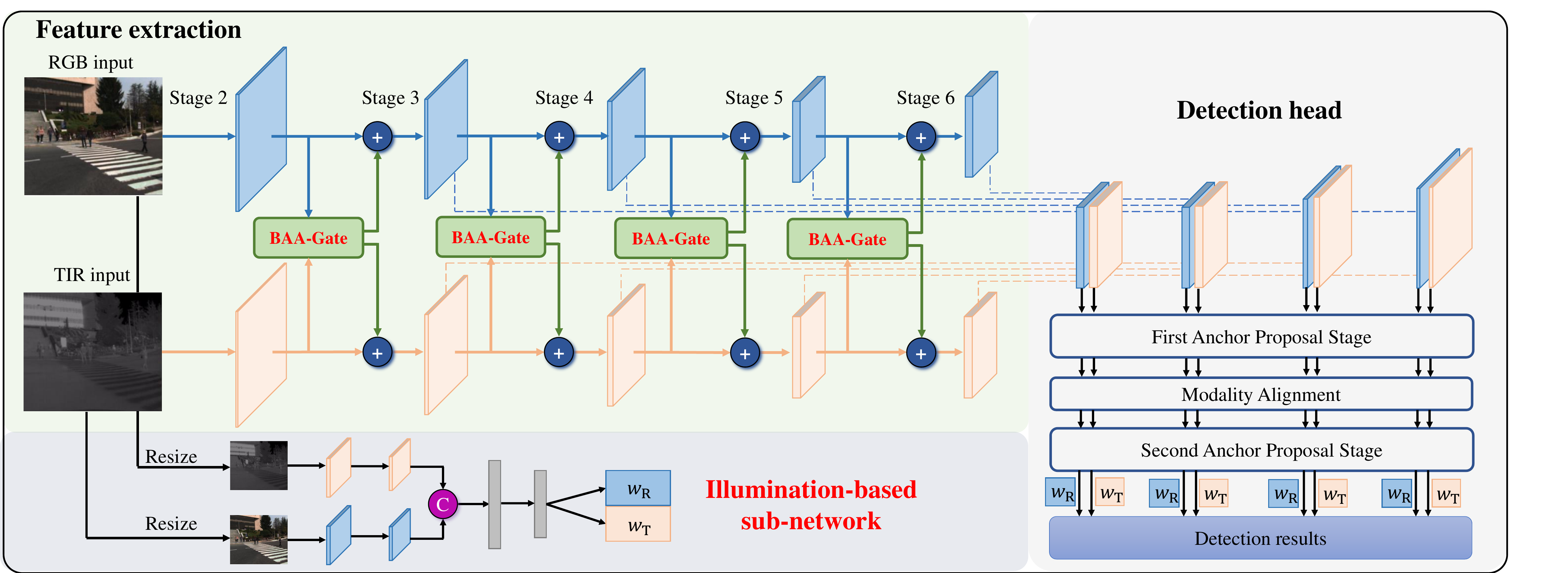}
	\caption{The overall architecture of the proposed BAANet.
		The BAANet consists of three parts, \textit{i.e.}, the feature extraction part, the illumination-based sub-network, and the detection head.
		The BAA-Gates embedded in the feature extraction framework distills and recalibrates  the modalities in a bi-direction manner via multiple stages.
		The illumination-based sub-network is proposed to adaptively adjust the interaction strength between modalities of the BAA-Gate and enhance the robustness towards illumination changes. 
	}
	\label{fig:mainstructure}
	\end{figure*}

	\section{RELATED WORKS}\label{sec:RELATEDWORK}

	\subsection{Multispectral pedestrian detection}
	Since the KAIST multispectral pedestrian dataset~\cite{hwang2015multispectral} proposal, the algorithms based on multispectral data have emerged rapidly~\cite{liu2016multispectral, choi2016multi, konig2017fully, guan2019fusion, zhang2020multispectral, li2019illumination, zhang2019weakly, zhang2021deep}. 
	Hwang \textit{et al}.~\cite{hwang2015multispectral} extend the ACF~\cite{Dollar2014ACF} approach leveraging the color-thermal image pairs as the initial baseline.
	Liu \textit{et al}.~\cite{liu2016multispectral} design four CNN based fusion architectures which merge two sub-networks on various DNNs stage.
	Li \textit{et al}.\cite{li2019illumination} introduce an illumination-based weighting mechanism for merging the color and thermal sub-networks to obtain the final confidence scores.
	CIAN~\cite{zhang2019cross} proposes a cross-modality interactive attention mechanism to leverage the modal correlations and adaptively fuse features.
	To tackle the position shift problem in color-thermal image pairs, AR-CNN~\cite{zhang2019weakly} designs a Region Feature Alignment module to align the region features of two modalities.
	Generally, these traditional methods are based on the simple two-branch architecture by concatenating the feature maps directly.
	However, we argue that these methods limit the representation potential due to the implicit modeling of the correlation between modalities.
	While our BAANet proposes the BAA-Gate to exploit the correlation between modalities and focus on distilling and recalibrating two modalities to improve the quality of representations effectively.
	
	\subsection{Attention mechanism} 
	The idea of attention mechanisms, which is inspired by the human visual attention mechanisms, is widely applied in various computer vision tasks, including image classification~\cite{Wang2017CVPR, hu2018squeeze}, semantic segmentation~\cite{ren2017end, fu2019dual}, image captioning~\cite{xu2015show, Huang_2019_ICCV}, \textit{etc}.
	For example, SENet~\cite{hu2018squeeze} recalibrates channel-wise feature maps by exploiting the correlations among channels.
	SKNet~\cite{Li2019CVPR} designs Selective Kernel units to adaptively fuse branches with different kernel sizes depending on the input information.
	CBAM~\cite{Woo2018ECCV} computes the attention maps along the channel and spatial dimension sequentially and refines the features adaptively.  
	For the RGB-T pedestrian detection, the core challenge is how to fully utilize both modalities in case of the essential difference between them and modality-specific noise in both modalities.
	In our work, the channel-wise and spatial-wise attention mechanisms are tailored and devised as the BAA-Gate.
	The BAA-Gate specializes in distilling the noise and spotlighting the complementary features to refine the counterpart modality further to improve unified representations.  
	

	\section{PROPOSED DETECTING APPROACH}\label{sec:METHOD}
In this section, the proposed BAANet is introduced in detail. 	 
Concretely, the design of BAA-Gate is introduced in subsection~\ref{subsec:CPFBlock}.
The illumination-based weighting strategy is introduced in subsection~\ref{sec:Illuminationstrategy}.
Finally, the architecture of the BAANet is introduced in subsection~\ref{sec:overallarchi}, which is illustrated in Fig.~\ref{fig:mainstructure}.

\subsection{Bi-directional Adaptive Attention Gate}\label{subsec:CPFBlock}
\begin{figure*}[!h]
	\centering
	\includegraphics[width=1\textwidth]{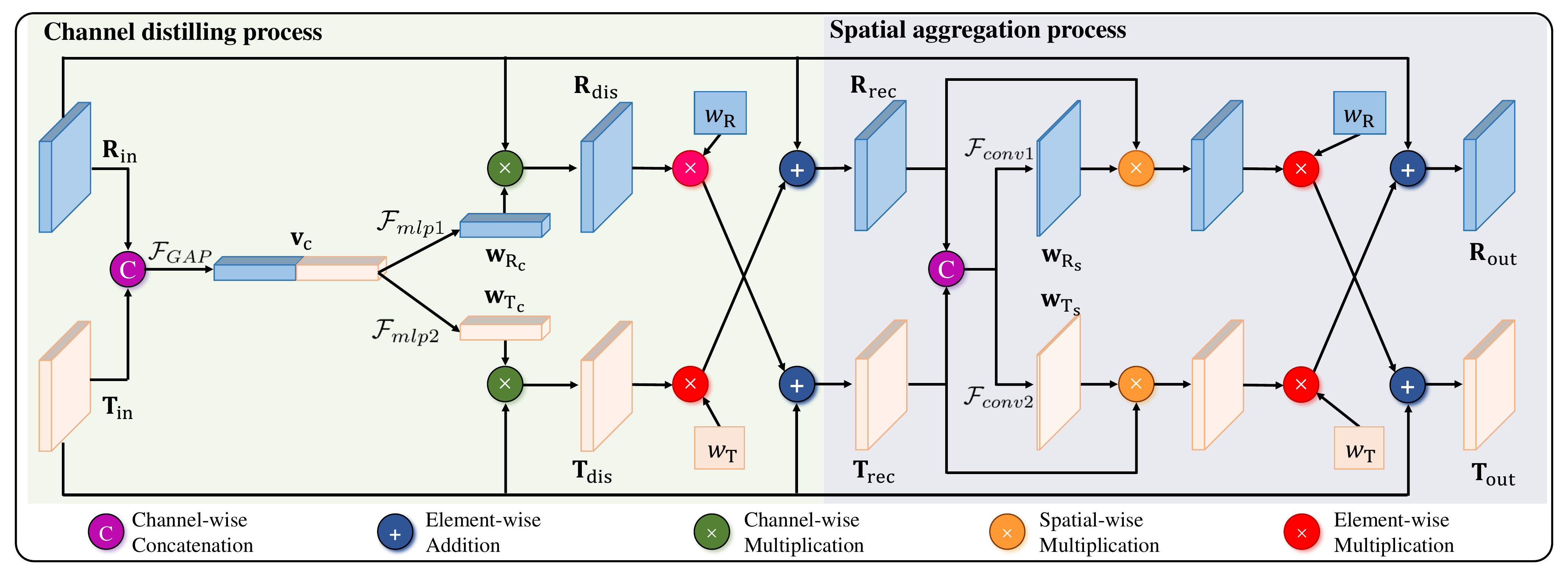}
	\caption{The detailed architecture of the designed Bi-directional Adaptive Attention Gate (BAA-Gate).
		The BAA-Gate is composed of two processes, \textit{i.e.}, the channel distilling process and spatial aggregation process.
		Inspired by the attention mechanisms, the channel distilling is devised to suppress noise and conduct recalibration on each single modality, and the spatial aggregation process is for aggregating the effective representations.
		Moreover, the adaptive interaction strength is introduced by the illumination-based weights to improve the robustness for illumination variations.
	}
	\label{fig:CFB}
\end{figure*}

The effective integration of RGB and TIR modalities is essential for RGB-T pedestrian detection.
However, the unavoidable noises exist naturally in two modalities, which bring great challenges to the fusion process.
Concretely, the TIR noise reflects as the objects with similar temperatures are prone to mix with surroundings.	
In contrast, the RGB noise is usually caused by complex lighting conditions and leads to similar object appearances, which are inscrutable.
To suppress the noise and fully exploit the inherent correlation between modalities, our work proposes the Bi-directional Adaptive Attention Gate (BAA-Gate).
Inspired by the attention mechanism, the proposed BAA-Gate is designed as two processes, \textit{i.e.}, the channel distilling and the spatial aggregation process, respectively.
The channel distilling is devised to suppress noise and conduct recalibration on each single modality, and the spatial aggregation process is for aggregating the effective representations, which are shown in Fig.~\ref{fig:CFB}.

\subsubsection{Channel distilling process}
Take the TIR modality as an example.
Due to the imaging mechanism of TIR camera, noise often reflects as objects with similar temperatures in local areas have a similar appearance in the TIR domain, which could lead to confusing features. 
Therefore, before the feature aggregation, the framework is desired to suppress the confusing TIR noise in the neighborhood of the object to prevent the bewildering noise from interfering with the recalibration of the RGB modality and features aggregation.
To achieve this, the channel distilling process explores to utilize the high confidence activations of RGB modality to suppress the corresponding low confidence TIR activations containing undesired noise.
In practice, the global spatial information of two modalities is first compressed to obtain a channel attention vector.
Concretely, given the input RGB and TIR modality $\mathbf{R}_{\rm{in}}$ and $\mathbf{T}_{\rm{in}}$, a channel attention vector ${\mathbf{v}}_{\rm{c}} \in \mathbb{R}^{2C \times 1 \times 1}$ is first obtained as follows:
\begin{equation}\label{12312312}
\mathbf{v}_{\rm{c}} = \mathcal{F}_{GAP}{(\mathbf{R}_{\rm{in}} \oplus \mathbf{T}_{\rm{in}})} \ ,
\end{equation}
where $\mathcal{F}_{GAP}$ denotes the global average pooling operation.
$\mathbf{R}_{\rm{in}}\in \mathbb{R}^{C\times H\times W}$ and $\mathbf{T}_{\rm{in}} \in \mathbb{R}^{C \times H \times W}$ denote the RGB and TIR input features. $\oplus$ denotes the concatenation.
The channel attention vector ${\mathbf{v}}_{\rm{c}}$ is a cross-modality representation that gives a global description about the informative channels of all input features. 
Then, through the multilayer perceptrons, the channel distilling weights ${\mathbf{w}_{\rm{T_{c}}}} \in \mathbb{R}^{C \times 1 \times 1}$ can be learned as follows:
\begin{equation}\label{123}
\mathbf{w}_{\rm{T_c}} = \sigma (\mathcal{F}_{mlp2}(\mathbf{v}_c)) \ ,
\end{equation}
where $\mathcal{F}_{mlp2}$ denotes the multilayer perceptrons, and $\sigma $ indicates the sigmoid excitation function to normalize the output in the range of (0, 1).
In this way, the channel distilling weights can learn from the correlation between modalities, therefore suppressing the exceptional TIR noise and enhance the strength of informative features.
Thus, the distilled TIR features can be highlighted by the channel-wise multiplication between the weights and the input features as:
\begin{equation}
\mathbf{T}_{\rm{{dis}}} = \mathbf{w}_{\rm{T_c}} \odot \mathbf{T}_{\rm{in}} \ ,
\end{equation}
where $\mathbf{T}_{\rm{{dis}}} \in \mathbb{R}^{C \times H \times W}$ denotes the distilled features. $\odot$ denotes the channel-wise multiplication operation.
Therefore, we reduce the noise level in the distilled features to provide more accurate TIR features for the recalibration of the RGB modality.
Then, the recalibration of RGB modality can be formulated as the summation between the RGB input features $\rm{\mathbf{R}_{in}}$ and the distilled TIR features $\mathbf{T}_{\rm{dis}}$ as follows:
\begin{equation}
\rm{\mathbf{R}_{rec}} = \rm{\mathbf{R}_{in}} +  \mathbf{T}_{\rm{dis}} \ ,
\end{equation}
where $+$ indicates the element-wise summation.
By doing so, the informative members in the distilled TIR features $\mathbf{T}_{\rm{dis}}$ can provide a recalibration in a certain position of the RGB modality.

Moreover, to enhance the robustness towards illumination changes, the illumination-based weighting strategy is introduced to achieve the adaptive interaction strength between modalities.
The illumination-based weight $\mathit{w}_{\rm{T}}$ can be introduced as follows:
\begin{equation}
\rm{\mathbf{R}_{rec}} = \rm{\mathbf{R}_{in}} + \mathit{w}_{\rm{T}} * \mathbf{T}_{\rm{dis}} \ ,
\end{equation}
where the ${w}_{\rm{T}}$ denotes the illumination-based weight, which reflects the confidence level of the TIR modality.
In this way, the RGB modality can be adaptively recalibrated by the TIR features with high confidence and avoid distraction from the low-quality features.
Note that the production of the illumination-based weight will be discussed in detail in the following subsection.

Meanwhile, since the RGB modality is also noisy, the bi-directional manner is devised to distill the RGB stream similarly and refine the TIR modality for the following feature aggregation. 

\subsubsection{Spatial aggregation process}
Considering the RGB and TIR modality are highly complementary, the network is expected to aggregate the two modalities complementarily at a certain position of the image according to their essential characteristics.
To do this, our work proposes utilizing the spatial attention mechanism to obtain the aggregation gates to reweight the activation intensity at different spatial positions of the RGB and TIR features.
In practice, the recalibrated RGB features $\rm{\mathbf{R}_{rec}}$ and TIR features $\rm{\mathbf{T}_{rec}}$ are chosen to generate preciser aggregation gates.
First, the input features of two modalities are concatenated to align them in a spatial-wise manner.
Then, a convolution function is adopted to learn the spatial aggregation gates from the recalibrated features.  
The spatial aggregation gate can be obtained as follows:
\begin{equation}
\mathbf{w}_{\rm{T_{s}}} = \mathcal{F}_{conv2}(\rm{\mathbf{R}_{rec}} \oplus \mathbf{T}_{\rm{rec}} ) \ ,
\end{equation}
where $\mathcal{F}_{conv2}$ denotes the convolution operation, which is implemented by the $1\times 1$ convolution operation.    $\mathbf{w}_{\rm{T_{s}}} \in \mathbb{R}^{ 1 \times H \times W}$ denotes the spatial aggregation gate. 
Further, each position of the TIR features is reweighted by the corresponding element in the spatial aggregation gate. 
$\rm{\mathbf{T}_{rec}}$ is also adjusted by the illumination-based weight.
Then, we obtain the final features $\rm{\mathbf{R}_{{out}}}$ for the spatial dimension as follows:
\begin{equation}
\rm{\mathbf{R}_{out}} = \rm{\mathbf{R}_{rec}} + \mathit{w}_{\rm{T}} *( \mathbf{w}_{\rm{T_{s}}}  \odot \mathbf{T}_{\rm{rec}} ) \ .
\end{equation}
where $\odot$ denotes the spatial-wise multiplication.
Similarly, the output features for TIR modality are aggregated as $\rm{\mathbf{T}_{out}}$ in the same manner.

So far, we obtain the refined output features $\mathbf{R}_{\rm{out}}$ and $\mathbf{T}_{\rm{out}}$ by cross-modality distilling and recalibration.
Then, the output features will be transported to the next stage to further refine the multi-modality features via multiple stages.

\subsection{Illumination-based weighting strategy }\label{sec:Illuminationstrategy}
Due to the RGB and the TIR modalities specialize in capturing different spectral, two modalities perform differently under various illumination conditions.
Hence, an illumination-based weighting strategy is designed to achieve adaptive interaction strength and enhance robustness in case of illumination changes.
Concretely, an illumination-based sub-network is proposed to determine the illumination condition, which is shown in Fig.~\ref{fig:mainstructure}.
To reduce the computational burden, input images are first resized to the size of $56 \times 56$.
Then, after two convolution layers, the RGB and TIR features are concatenated, and two fully connected layers and a softmax layer are adopted to calculate the probability of day and night.

\begin{figure}[!h]
	
	\includegraphics[width=0.45\textwidth]{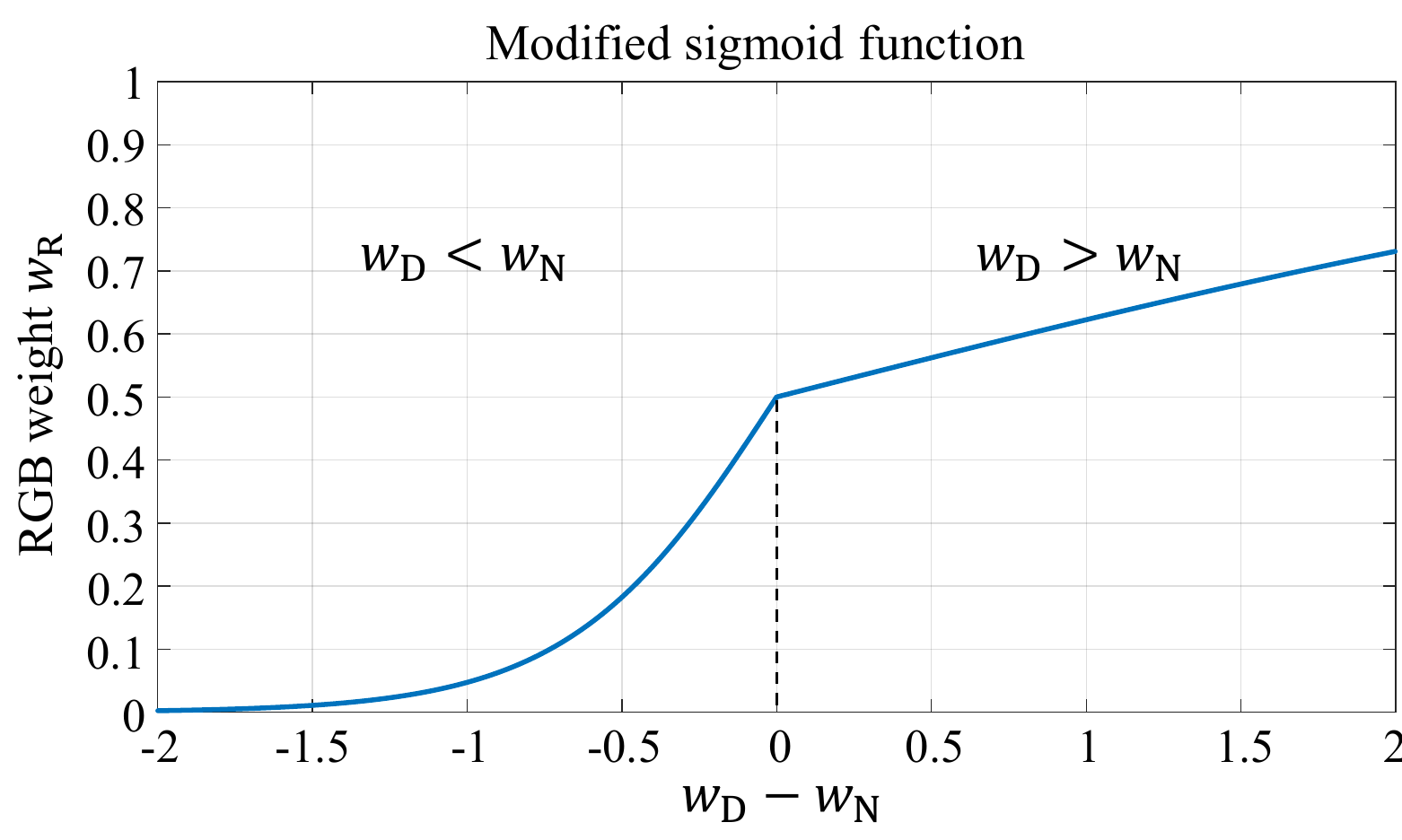}
	\caption{
		The illustration of the modified sigmoid function.  
   		When the probability of night is greater than that of daytime, the weight of RGB modality drops sharply to avoid distraction due to poor illumination.
   		While when the probability of daytime is greater, thermal features still occupy a large weight to assist with the RGB features for detection. 
	}
	\label{fig:modified_simoid}
\end{figure}

For training, the captured time (day or night) is introduced as the ground truths.
The loss function is formulated as:
\begin{equation}
L_{I} = -\hat{w}_{\rm{D}} \cdot log(w_{\rm{D}}) +  -\hat{w}_{\rm{N}} \cdot log(w_{\rm{N}}) \ ,
\end{equation}
where $\hat{w}_{\rm{D}}$ and $\hat{w}_{\rm{N}}$ denote the ground truths of day and night. ${w}_{\rm{D}}$ and ${w}_{\rm{N}}$ denote the probability of the daytime and nighttime.
Moreover, for better balancing the contribution of two modalities, RGB and TIR weights $w_{\rm{R}}$ and $w_{\rm{T}}$ are obtained by a modified sigmoid function as follows:
\begin{equation}
\left\{
\begin{array}{l}
w_{\rm{R}} = 
\left\{
\begin{array}{rcl}
\frac{1}{1 + e^{- k_1 * (w_{\rm{D}}- w_{\rm{N}})}}
& {w_{\rm{D}} > w_{\rm{N}}}\\
\frac{1}{1 + e^{- k_2 * (w_{\rm{D}}- w_{\rm{N}})}}     & {w_{\rm{D}} < w_{\rm{N}}}
\end{array} 
\right. \\
w_{\rm{T}} = 1-w_{\rm{R}}

\end{array}\ ,
\right.
\end{equation}
where $k_1$ and $k_2$ are the hyperparameters to modified the sigmoid function. 
The illustration of the function is shown in Fig.~\ref{fig:modified_simoid}, where $k_1$ and $k_2$ are set to 0.5 and 3.
Due to the TIR features are robust towards the illumination changes, the TIR weight $w_{\rm{T}}$  contribute equally to the detection in day and night.
Considering the RGB modality are prone to introduce unexpected noise at night, the RGB weight $w_{\rm{R}}$ is reduced rapidly in poor illumination conditions. 
Therefore, the interaction strength between modalities can be adaptively adjusted according to the illumination conditions, significantly enhances the robustness towards complex illumination.

\subsection{Overall architecture}\label{sec:overallarchi}
The architecture of BAANet includes three parts: the feature extraction, the illumination-based sub-network and the detection head.
For the feature extraction part, the proposed BAA is embedded in the ResNet-50 as shown in Fig.~\ref{fig:mainstructure}.
For the detection head, inspired by ALFNet~\cite{Liu_2018_ECCV}, the cascade-predictors strategy is introduced to improve the detection performance progressively.
Then, the Deformable Convolution Networks (DCN)~\cite{dai2017deformable} are introduced as the modality alignment module to relieve the weak alignment problem between two modalities.
Meanwhile, the illumination-based weights are introduced to reweight the confidence scores.
Therefore, the cascade detection process can be formulated as follows:
\begin{equation}
\left\{
\begin{array}{l}
c_{final} = c_1 \cdot c_2 = c_1 \cdot (w_{\rm{R}} \cdot c_{\rm{R}} + w_{\rm{T}} \cdot c_{\rm{T}})\\
b_{final} = b_1 + b_2
\end{array}
\right . \  , 
\end{equation} 
where $c_1$ and $c_2$ denote the confidence scores predicted from the first and the second AP stage respectively.
$c_{\rm{R}}$ and $c_{\rm{T}}$ denote the confidence scores predicted from each modality.
$b_1$ and $b_2$ denote the predicted bounding boxes regression offsets from two stages, respectively.
By the summation of $b_1$ and $b_2$, the bounding boxes can be progressively evolved and asymptotically located.
Inspired by~\cite{lin2017focal}, the focal weight is adopted to relieve the positive-negative imbalance problem.
The classification loss $L_{cls}$ can be formulated as follows:
\begin{equation}
L_{cls} = -\alpha \sum_{i \in S_+}(1 - c_i)^{\gamma} log(c_i) -(1-\alpha)\sum_{i \in S_{-}}c_i^{\gamma} log(1-c_i) \ ,
\end{equation}
where $S_+$ and $S_-$ denote the positive and negative anchor boxes.
$c_i$ denotes the confidence score of sample $i$.
$\alpha$ and $\gamma$ denote the focusing parameters, which are experimentally set as $\alpha = 0.25$ and $\gamma = 2$ suggested in~\cite{lin2017focal}.
The total loss is the sum of classification loss $L_{cls}$, regression loss $L_{reg}$ and the illumination loss $L_{I}$.
The smooth L1 loss proposed by \cite{Ren2017Faster} is adopted as the regression loss $L_{reg}$.
The total loss function is formulated as follows:
\begin{equation}
L = L_I + L_{cls1} + L_{cls2} + [y=1]L_{reg1} + [y=1]L_{reg2}    \ .
\end{equation}

Based on the BBA-Gates embedded in the backbone, the RGB and TIR modality can be both distilled and recalibrated for comprehensive representations gradually in a bi-directional manner.
Through the cascade detections, the bounding boxes are progressively refined for more precise localization.
Meanwhile, the illumination-based strategy also strikes a balance between the RGB and the thermal modalities, which further improves the adaptivity for challenging illumination changes.
    
\begin{table*}[htbp]
	\centering
	\caption{Miss rate comparison on all nine subsets of the KAIST~\cite{hwang2015multispectral} dataset.
		Sca. and Occ. denote the scale and the occlusion, respectively.
		\textcolor[rgb]{1,0,0}{Red}, \textcolor[rgb]{0,1,0}{green} and \textcolor[rgb]{0,0,1}{blue} color denote the
		first, second and third place respectively.
	}
	\setlength{\tabcolsep}{2mm}{

		\begin{tabular}{ccccccccccccc}
			\toprule
			\toprule
			Methods & All & Day & Night & Sca.-Near & Sca.-Medium & Sca.-Far & Occ.-None & Occ.-Partial & Occ.-Heavy   & Year \\
			\midrule
			ACF~\cite{hwang2015multispectral}   & 47.32\%  & 42.57\%  & 56.17\%  & 28.74\%  & 53.67\%  & 88.20\%  & 62.94\%  & 81.40\%  & 88.08\%     & 2015 \\
			\midrule
			Halfway Fusion~\cite{liu2016multispectral} & 25.75\%  & 24.88\%  & 26.59\%  & 8.13\%  & 30.34\%  & 75.70\%  & 43.13\%  & 65.21\%  & 74.36\%    & 2016 \\
			\midrule
			Fusion RPN+BF~\cite{konig2017fully} & 18.29\%  & 19.57\%  & 16.27\%  & 0.04\%  & 30.87\%  & 88.86\%  & 47.45\%  & 56.10\% & 72.20\%     & 2017 \\
			\midrule
			MSDS-RCNN~\cite{li2018multispectral} & 11.63\%  & 10.60\%  & 13.73\%  & 1.29\%  & 16.19\%  & 63.73\%  & \textcolor{blue}{29.86\%}  & 38.71\%  & 63.37\%     & 2018 \\
			\midrule
			IAF R-CNN~\cite{li2019illumination} & 15.73\%  & 14.55\%  & 18.26\%  & 0.96\%  & 25.54\%  & 77.84\%  & 40.17\%  & 48.40\%  & 69.76\%     & 2019 \\
			\midrule
			IATDNN + IASS~\cite{guan2019fusion} & 14.95\%  & 14.67\%  & 15.72\%  & 0.04\%  & 28.55\%  & 83.42\%  & 45.43\%  & 46.25\%  & 64.57\%     & 2019 \\
			\midrule
			CIAN~\cite{zhang2019cross}  & 14.12\%  & 14.77\%  & 11.13\%  & 3.71\%  & 19.04\%  & \textcolor{green}{55.82\%}  & {30.31\%}  & 41.57\%  & 62.48\%    & 2019 \\
			\midrule
			AR-CNN~\cite{zhang2019weakly} & \textcolor{blue}{9.34\%}  & 9.94\%  & \textcolor{blue}{8.38\%}  & \textcolor{blue}{0.00\%}  & \textcolor{blue}{16.08\%}  & 69.00\%  & 31.40\%  & \textcolor{blue}{38.63\%}  & \textcolor{red}{55.73\%}     & 2019 \\
			\midrule
			CFR~\cite{zhang2020multispectral}   & 10.05\%  & \textcolor{blue}{9.72\%}  & 10.80\%  & -     & -     & -     & -     & -     & -        & 2020 \\
			\midrule
			MBNet~\cite{MBNetECCV2020} & \textcolor{green}{8.13\%} & \textcolor{red}{8.28\%} & \textcolor{green}{7.86\%}  & \textcolor{green}{0.00\%}  & \textcolor{green}{16.07\%}  & \textcolor{blue}{55.99\%}  & \textcolor{green}{27.74\%}  & \textcolor{green}{35.43\%} & \textcolor{blue}{59.14\%}   & 2020 \\
			\midrule
			{BAANet(Ours)} & \textcolor{red}{7.92\%}  & \textcolor{green}{8.37\%}  & \textcolor{red}{6.98\%} & \textcolor{red}{0.00\%} & \textcolor{red}{13.72\%} & \textcolor{red}{51.25\%} & \textcolor{red}{25.15\%} & \textcolor{red}{34.07\%}  & \textcolor{green}{57.92\%}  & 2021 \\
			\bottomrule
			\bottomrule
		\end{tabular}%
	}
	\label{tab:overallcomparison}%
\end{table*}%

	\section{EXPERIMENTS}\label{sec:EXPERIMENT}
In this section, we first introduce the challenging KAIST multispectral pedestrian detection dataset~\cite{hwang2015multispectral}.
Then, we show the implementation details and the experimental results.
Finally, the ablation experiments are conducted to verify the effectiveness of the proposed components.

\subsection{KAIST multispectral pedestrian Dataset}
The KAIST dataset~\cite{hwang2015multispectral} consists of 95,328 aligned RGB-T image pairs under different illumination conditions.
The annotations cover 1,182 pedestrians with a total of 103,128 bounding boxes.
Considering the faulty annotations in the training dataset, the improved annotations proposed by~\cite{zhang2019weakly} are adopted for training.
The testing set consists of 2,252 image pairs, in which 1,455 images during the daytime and 797 images at nighttime.
Moreover, we adopt the test annotations improved by \cite{liu2016multispectral} for the evaluation process.




\subsection{Implementation details}
\textbf{Platform.} The BAANet is implemented on a computer with 16GB RAM, Xeon(R) E5-2689(2.60GHz) CPU and a single Nvidia GeForce GTX 1080Ti GPU.


\textbf{Parameters setting.} The proposed BAANet is trained for eight epochs utilizing the Adam optimizer with a batch size of 8 and a learning rate of 0.0001.
The training IoU of the first and the second anchor proposal stage is set to \{0.3, 0.5\} and \{0.5, 0.7\}, respectively.
The Xavier approach~\cite{glorot2010understanding} is adopted for the random initialization of the convolution layers.
The anchor ratio is set to 0.41.


\textbf{Evaluation metric.} The log-average Miss Rate (MR$^{-2}$) over False Positive Per Image (FPPI) range of [10$^{-2}$, 10$^0$] is employed for the evaluation, which is consistent with the original KAIST evaluation metric.
Besides, the reasonable setup, \textit{i.e.}, not or partially occluded pedestrians which are larger than 55 pixels, is adopted.

\begin{figure}[t]
	\centering
	\includegraphics[width=0.44\textwidth]{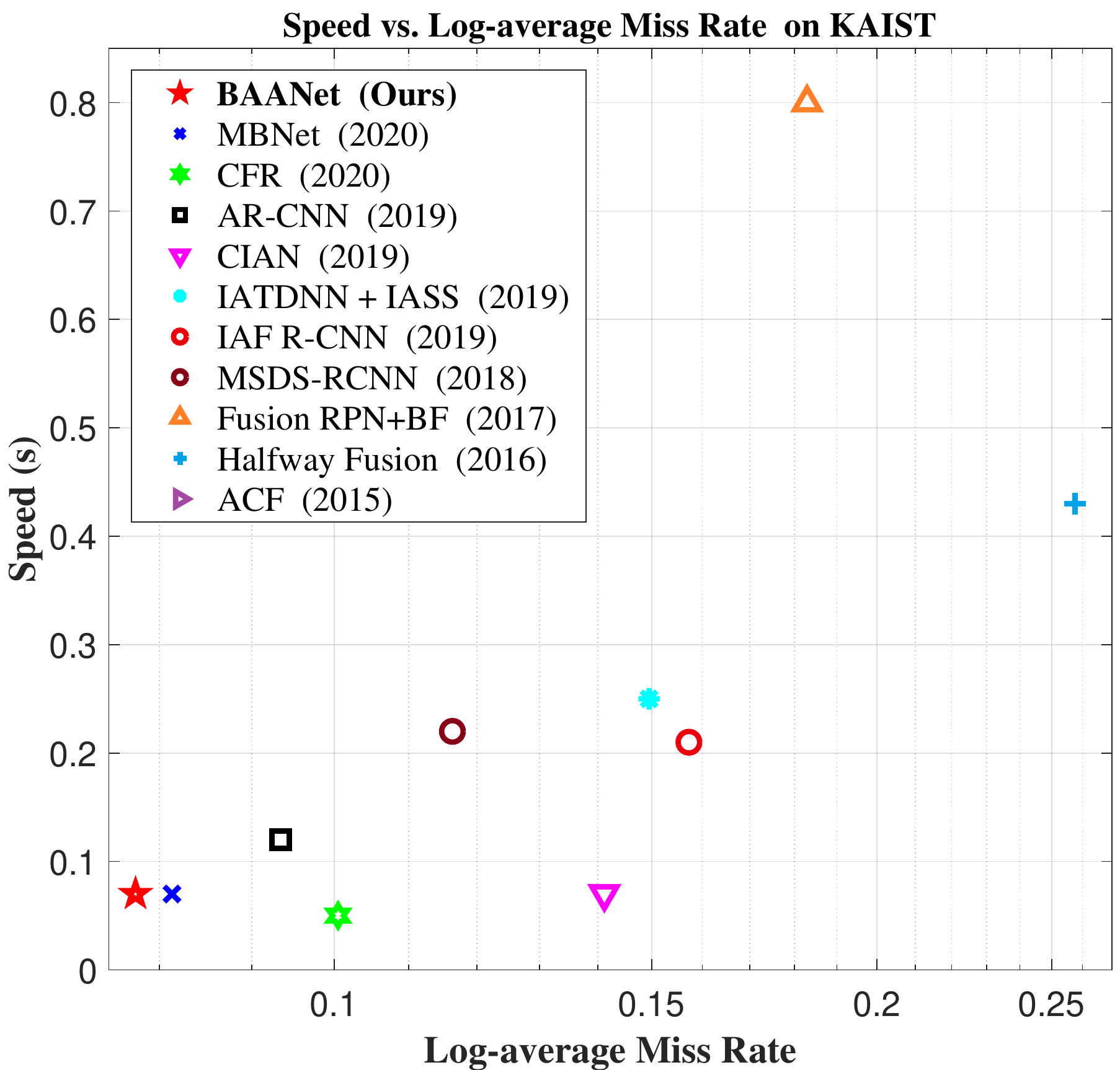}
	\caption{Log-average Miss rate versus the speed of the proposed BAANet detector.
		It can be seen that the proposed BAANet performs outstandingly in both running speed and miss rate. 	
	}
	\label{fig:stplot}
	\vspace{-10pt}
\end{figure}

\begin{figure}[t]
	\centering
	\includegraphics[width=0.48\textwidth]{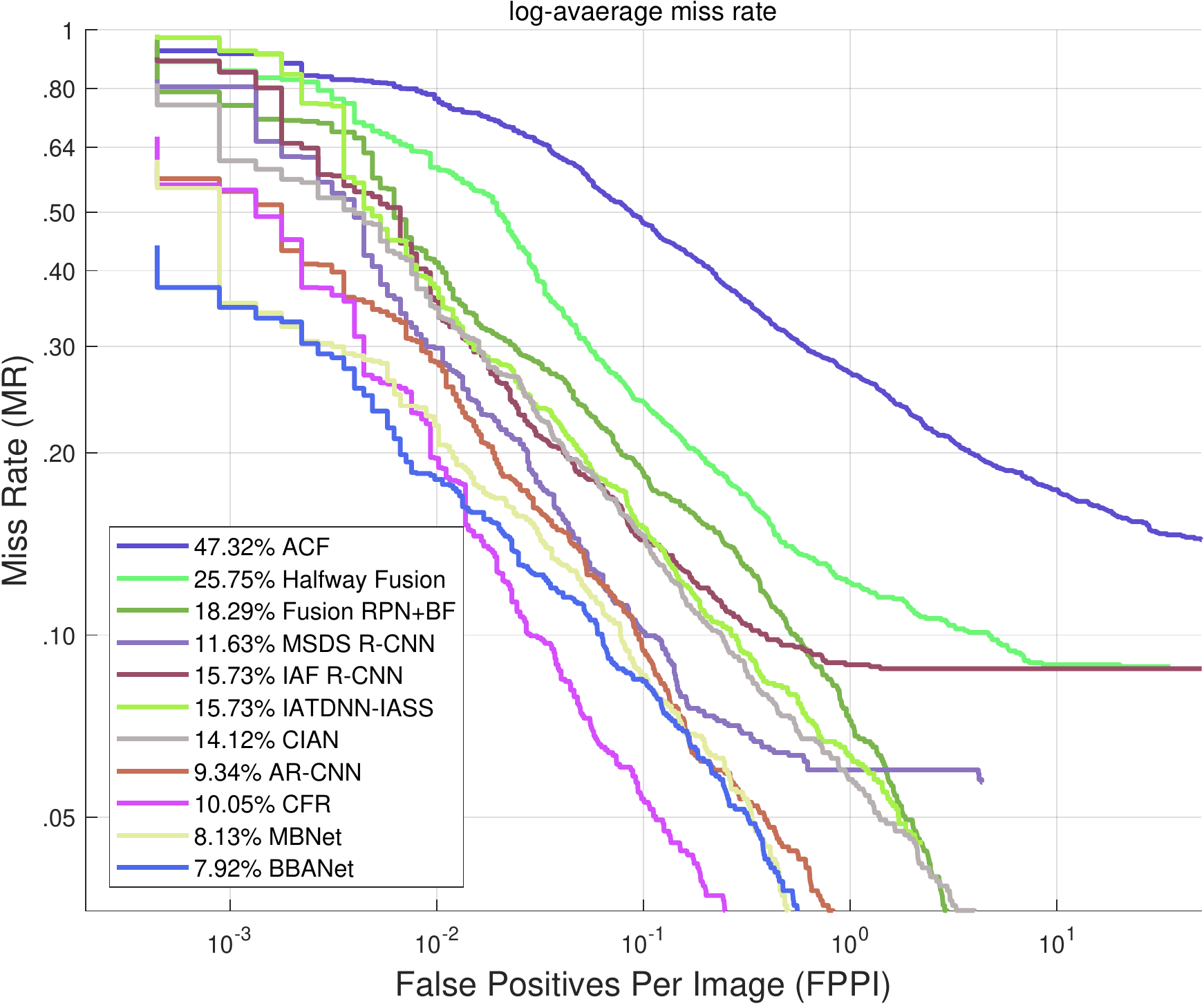}
	\caption{Comparison of the detection results with the state-of-the-art methods on the KAIST~\cite{hwang2015multispectral} dataset under the reasonable subset.		
 }
		\vspace{-15pt}
	\label{fig:FPPI-MR}
\end{figure}

\begin{figure*}[!h]
	\includegraphics[width=1\textwidth]{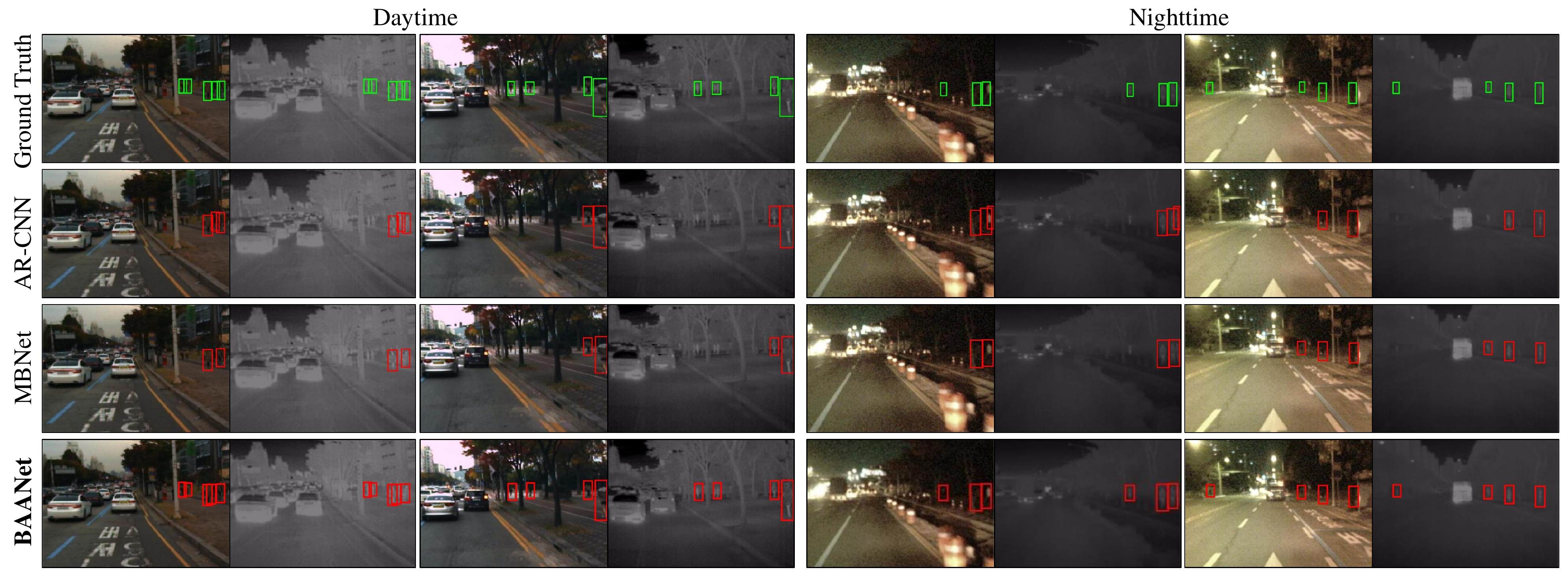}
	\caption{Qualitative evaluation of BAANet with two top ranked methods, \textit{i.e.}, MBNet~\cite{MBNetECCV2020} and AR-CNN~\cite{zhang2019weakly}.
		The green rectangles denote the ground truth, while the red ones denote the detection results.
	}
	\label{fig:Qualitative evaluation}
		\vspace{-10pt}
\end{figure*}

\subsection{Evaluation on the KAIST dataset}
\label{subsec:overall}
\textbf{Quantitative evaluation.} 
The BAANet detector is compared with other state-of-the-art methods in nine subsets of the KAIST dataset, \textit{i.e.}, all, day and night subsets
of the captured time, near, medium and far subsets of the pedestrian scale, and none, partial and heavy subsets of the occlusion levels.
As shown in Table~\ref{tab:overallcomparison}, the BAANet ranks first in overall performance on KAIST.
Besides, the BAANet rank first in seven of nine subsets. 
For instance, in the night subset, the proposed BAANet (6.98\%) surpasses the second-best (7.86\%) and the third-best (8.38\%) by 0.88\% and 1.4\% respectively, which indicates the satisfactory accuracy of BAANet at nighttime.
It can be attributed to the proposal of the BAA-Gate, the noise brought by the RGB modality in the night is suppressed and recalibrated, and the TIR features are aggregated to form the high-quality features.
Besides, the BAANet also takes the lead in all three subsets of the pedestrian scale, especially in the far subset, the BAANet (51.25\%) outperforms the second-best CIAN (55.82\%) and the third-best MBNet (55.99\%) by 4.57\% and 4.74\% respectively, which indicates that our BAANet has the remarkable ability for small scale object detection.
The log-average Miss Rate (MR$^{-2}$) over False Positive Per Image (FPPI) curve is shown in Fig.~\ref{fig:FPPI-MR}.

\textbf{Qualitative evaluation.} 
The detecting examples of our BAANet with other detectors are shown in Fig. \ref{fig:Qualitative evaluation}.
It can be seen that BAANet performs outstandingly in both day and night.
In case of poor illumination or small-scale pedestrians, the BAANet still locates the pedestrians accurately and robustly. 
While other detectors are prone to missed and false detection in challenging scenarios.

\textbf{Speed comparison.} 
The speed comparison of BAANet and other state-of-the-art methods is shown in Tabel \ref{tab:speedcom}.
Compared with other detectors at the same level of miss rate, the speed of BAANet still performs satisfactorily.
The speed versus miss rate of all methods is illustrated in Fig.~\ref{fig:stplot}, which shows the proposed methods performs remarkably in balancing miss rate and speed.

\begin{table}[htbp]
	\centering
	\caption{The speed comparison between the BAANet and other state-of-the-art multispectral detectors. }
		\renewcommand\arraystretch{0.5}
	\begin{tabular}{cccc}
		\toprule
		\toprule
		Methods & Speed(s) & Year  & Platform \\
		\midrule
		ACF~\cite{hwang2015multispectral}   & 2.73  & 2015  & MATLAB \\
		\midrule
		Halfway Fusion~\cite{liu2016multispectral} & 0.43  & 2016  & TITAN X \\
		\midrule
		Fusion RPN+BF~\cite{konig2017fully} & 0.80  & 2017  & MATLAB \\
		\midrule
		MSDS-RCNN~\cite{li2018multispectral} & 0.22  & 2018  & TITAN X \\
		\midrule
		IAF R-CNN~\cite{li2019illumination} & 0.21  & 2019  & TITAN X \\
		\midrule
		IATDNN + IASS~\cite{guan2019fusion}  & 0.25  & 2019  & TITAN X \\
		\midrule
		CIAN~\cite{zhang2019cross}  & 0.07  & 2019  & 1080 Ti \\
		\midrule
		AR-CNN~\cite{zhang2019weakly} & 0.12  & 2019  & 1080 Ti \\
		\midrule
		CFR~\cite{zhang2020multispectral}   & 0.05  & 2020  & 1080 Ti \\
		\midrule
		MBNet~\cite{MBNetECCV2020} & 0.07  & 2020  & 1080 Ti \\
		\midrule
		\textbf{BAANet(Ours)} & 0.07  & 2021  & 1080 Ti \\
		\bottomrule
		\bottomrule
	\end{tabular}%
	  		\vspace{-10pt}
	\label{tab:speedcom}%
\end{table}%
%
%

\begin{table}[htbp]
	\centering
	\caption{The ablation experiments of the BAANet with different components on KAIST dataset.}
	\begin{tabular}{ccccccc}
		\toprule
		\toprule
		
		BA-Gate & Illu. & MA   & All   & Day   & Night \\
		
		\midrule
		
		&       &          & 10.62\% & 11.05\% & 9.42\% \\
		
		\midrule
		
		\checkmark &       &        & 9.44\% & 10.49\%  & 7.45\% \\
		
		\midrule
		
		\checkmark & \checkmark &       & 8.22\%   & 8.61\% & 7.66\% \\
		
%
		
		\midrule
		
		\checkmark & \checkmark &  \checkmark& \textbf{7.92}\%  & \textbf{8.37}\% & \textbf{6.98}\%  \\
		
		\bottomrule
		\bottomrule
	\end{tabular}%
	\label{tab:ablation}%
	     	\vspace{-15pt}
\end{table}%

\subsection{Ablation study}
\label{subsec:Ablationstudy}
In this subsection, ablation experiments are conducted on the KAIST dataset to demonstrate the effectiveness of each component proposed in the BAANet.
Table~\ref{tab:ablation} shows comparisons of five versions of the BAANet, where the baseline adopts the simple concatenation as the fusion strategy. 
'BA-Gate' denotes the Bi-directional Attention Gate without adaptive interaction strength.
'Illu' denotes the illumination-based weighting strategy for adaptively adjusting the interaction strength.
'MA' indicates the Modality Alignment implemented by DCN.
The experiments show that the final version with all three designed components outperforms other versions.
Results of the ablation study demonstrate the effectiveness of the proposed components.

\section{CONCLUSIONS}\label{sec:CONCLUSIONS}
In this work, the BAANet embedded with novel Bi-directional Adaptive Attention Gates (BAA-Gates) is proposed.
Based on the attention mechanism, the BAANet first distills the modality-specific noise in each modality, and then recalibrates the counterpart modality via a bi-directional multi-stage fusion strategy.
Besides, the adaptive interaction strength of BAA-Gate is implemented by an illumination-based weighting strategy to enhance the adaptability for the illumination changes.
The experiments demonstrate that the proposed BAANet performs outstandingly compared with the other ten detectors in both accuracy and speed.
Consequently, we hope that our work could contribute to the development of multispectral applications. 



\section*{ACKNOWLEDGMENT}
This work was supported by the National Natural Science Foundation of China (Grant No. 62103261), Postdoctoral Science Foundation of China (Grant No. 2020M681301), and Science and Technology on Near-Surface Detection Laboratory(Grant No. 6142414190203).
	
	{\small
		
		\bibliographystyle{IEEEtran}
		\bibliography{ICRA2022}
	}
	
\end{document}